%% file: main.tex
\documentclass[ sigconf, review=false, preprint]{acmart}

\usepackage{booktabs} %
\usepackage{hyperref}
\usepackage{algorithm}
\usepackage{algpseudocode}
\usepackage{tikz}
\usetikzlibrary{shapes, arrows, positioning}
\usepackage{subcaption}
\usepackage{booktabs}
\usepackage{multirow}
\usepackage{float}
\usepackage{xurl}
\usepackage{ragged2e}

\begin{document}
\title{Accelerating SfM-based Pose Estimation with Dominating Set}

\author{Joji Joseph}
\affiliation{%
  \institution{Indian Institute of Science}
  \city{Bangalore}
  \state{Karnataka}
  \country{India}
  \postcode{560012}
}

\author{Bharadwaj Amrutur}
\affiliation{%
  \institution{Indian Institute of Science}
  \city{Bangalore}
  \state{Karnataka}
  \country{India}
  \postcode{560012}
}

\author{Shalabh Bhatnagar}
\affiliation{%
  \institution{Indian Institute of Science}
  \city{Bangalore}
  \state{Karnataka}
  \country{India}
  \postcode{560012}
}

\renewcommand{\shortauthors}{}

\begin{teaserfigure}
    \justifying
    \begin{subfigure}[b]{0.48\textwidth}
    \frame{\includegraphics[width=\textwidth]{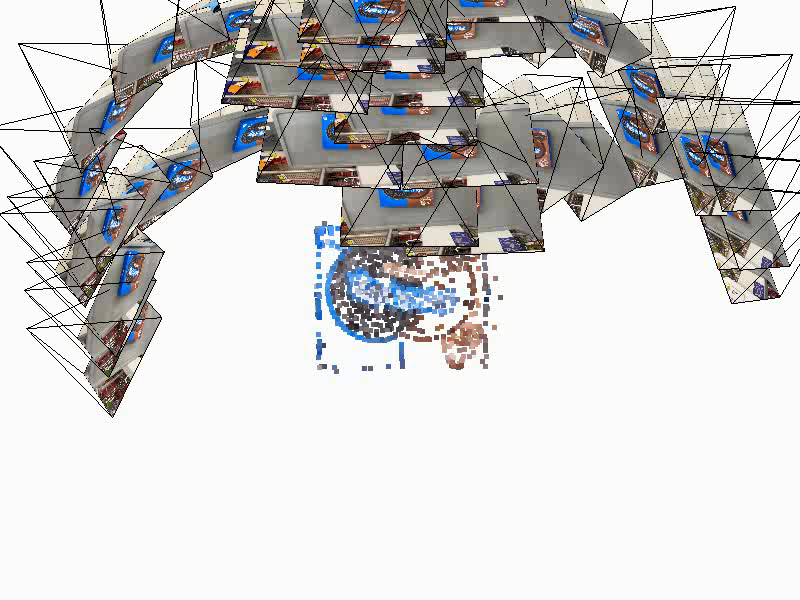}}
    \caption{}
    \end{subfigure} %
    \begin{subfigure}[b]{0.48\textwidth}
    \frame{\includegraphics[width=\textwidth]{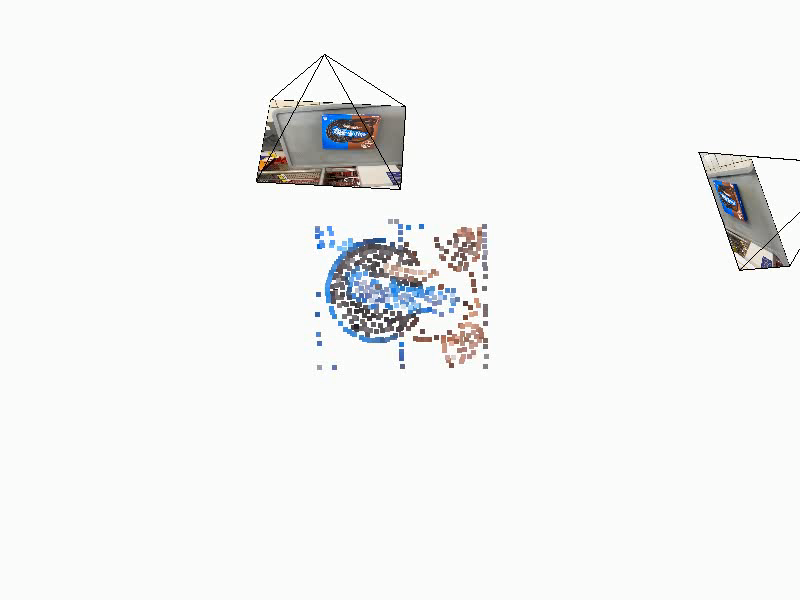}}
    \caption{}
    \end{subfigure}
    \caption{Impact of applying the dominating set on Structure from Motion (SfM). Each reference image is depicted along with the camera pose as a pyramid, with the apex representing the camera position and the base representing the image plane. (a) The SfM point cloud before applying the dominating set, illustrating the original density of reference images and points. (b) The SfM point cloud after applying the dominating set, showing a significant reduction in the number of reference images to two and a decrease in points, while retaining essential information necessary for accurate object pose estimation.}
    \label{fig:effect_of_dominating_set_on_sfm}
\end{teaserfigure}

\begin{abstract}
This paper introduces a preprocessing technique to speed up Structure-from-Motion (SfM) based pose estimation, which is critical for real-time applications like augmented reality (AR), virtual reality (VR), and robotics. Our method leverages the concept of a dominating set from graph theory to preprocess SfM models, significantly enhancing the speed of the pose estimation process without losing significant accuracy.  Using the OnePose dataset, we evaluated our method across various SfM-based pose estimation techniques.  The results demonstrate substantial improvements in processing speed, ranging from 1.5 to 14.48 times, and a reduction in reference images and point cloud size by factors of 17-23 and 2.27-4, respectively. This work offers a promising solution for efficient and accurate 3D pose estimation, balancing speed and accuracy in real-time applications. For more information, visit our project webpage at: {\color{blue} \url{accelerating-sfm-with-dominating-set.surge.sh}}.
\end{abstract}



\maketitle

\input{1_introduction}
\input{2_related}
\input{3_dominating_set}
\input{4_method}
\input{5_experiments}

\input{6_conclusion}

\bibliographystyle{ACM-Reference-Format}
\bibliography{references}

\end{document}

%% file: 1_introduction.tex
\section{Introduction}
Accurate pose estimation of novel objects is a crucial task in computer vision and robotics. This task involves precisely identifying an object's position and orientation, a key aspect in accurately manipulating objects or overlaying a virtual object in the input video stream for augmented reality. The challenge can be described simply as follows: Given a comprehensive 3D representation of an object, how do we pinpoint its pose in a given image? A comprehensive 3D representation allows for accurate pose estimation without relying on partial
models or single 2D images, enabling more reliable and precise determination
of an object’s pose.

There are various methods for representing objects, including collection of RGB images\cite{liu2022gen6d}, collection of RGB-D images\cite{he2022fs6d}, 3D models, and Structure-from-Motion (SfM)\cite{sun2022onepose, he2022oneposeplusplus}. Among these, SfM-based approaches, which use monocular camera images to reconstruct 3D structures, are noteworthy for their high accuracy. However, a significant drawback of SfM is its slower processing time compared to other methods. This slower speed is a bottleneck, especially in applications where quick decision-making is crucial.

Given the high accuracy of SfM-based methods, though they are generally slower, there is a compelling need to explore strategies to accelerate these methods. This is where our research introduces a novel approach. We propose optimizing the SfM representation using the concept of `dominating set' from graph theory. The idea of a dominating set, often used in network theory for efficient resource allocation and communication\cite{DomSetsBroadcasting_Stojmenovic2002, CDSWirelessSurvey_Yu2013}, can be applied to SfM to optimize the pose estimation speed.

The remainder of this paper is organized as follows: We begin by reviewing related works in SfM-based pose estimation (section~\ref{sec:prior}) to set the context for our contribution. Following this, we illustrate the concept of dominating sets from graph theory (section~\ref{sec:dominating_set}). Our novel approach of optimizing SfM using a minimum dominating set problem is presented thereafter (section~\ref{sec:sfm_with_dominating_sets}). Finally, we outline our experimental setup, discuss the results (section~\ref{sec:experiments}), and conclude with a summary of the paper (section~\ref{sec:conclusion}).

%% file: 2_related.tex
\section{Related Works}

\label{sec:prior}

\subsection{Prelims}
\subsubsection{PnP algorithm for 3D pose estimation}
The Perspective-n-Point (PnP) algorithm\cite{lepetit2009epnp} is pivotal in computer vision for estimating the pose of an object relative to a camera. Given the three-dimensional (3D) coordinates of an object within a specific coordinate frame and their corresponding projections onto an image plane, the PnP algorithm facilitates the determination of the object's pose.

With precisely three points, the algorithm is capable of producing up to four potential solutions due to inherent geometric ambiguities. The introduction of a fourth point enables the derivation of a unique solution, assuming noise-free and accurate correspondences. In practice, scenarios often involve more than four 2D-3D correspondences, necessitating robust statistical methods to mitigate the effects of outliers and measurement noise. A prevalent approach involves the application of the Random Sample Consensus (RANSAC) \cite{FischlerBolles1981RANSAC} methodology, wherein subsets of correspondences are randomly selected to estimate the object pose. This pose is then utilized to reproject all 3D coordinates back onto the image plane, facilitating the identification of inliers. Iterating this process allows for the selection of the solution that maximizes the number of inliers, thus ensuring a reliable estimation of the object pose.

\subsubsection{Feature Matching}
Feature matching is a crucial process in computer vision that involves identifying correspondences between points (features) across different images. These correspondences are vital not only for Perspective-n-Point (PnP) algorithms but also play a significant role in the broader context of 3D reconstruction.

Feature extraction algorithms\cite{Lowe2004DistinctiveIF,Bay2006SURF,detone2018superpoint,gleize2023silk} are designed to identify points that are invariant to changes in viewpoint, lighting, and other environmental factors. These algorithms construct descriptors that encode the contextual information of the points, facilitating their identification across different views.

Subsequently, feature-matching algorithms compare these descriptors to establish correspondences between features in different images. These algorithms range from simple methods based on cosine similarity to more advanced deep learning models, such as SuperGlue and LightGlue\cite{sarlin2020superglue, lindenberger2023lightglue}. Additionally, some approaches do not rely on extracted features, such as LoFTR\cite{sun2021loftr}, directly matching the points in two input images.

\subsubsection{Structure from Motion (SfM)}
Structure from Motion (SfM) is a complex process that leverages correspondences between features across multiple images to reconstruct a scene's 3D structure and determine the camera's motion or pose\cite{Hartley2004MultipleView, schoenberger2016sfm,schoenberger2016mvs}. By triangulating these matched features, SfM algorithms iteratively estimate both the camera poses and a sparse 3D model of the environment. This model is crucial for understanding the scene's geometry and facilitating the estimation of an object's pose within new viewpoints or environments.

To estimate the pose of a new viewpoint (or pose of the object), SfM systems employ feature-matching techniques that align 2D points from the novel viewpoint with the 3D points in the pre-constructed SfM model. Following this, Perspective-n-Point (PnP) algorithms utilize these 2D-3D correspondences to accurately estimate the pose of the camera or object in the new environment.

\subsection{SfM-based pose estimation}
The generalized SfM-based pose estimation pipeline is given in figure~\ref{fig:sfm_pose_estimation}.  Our preprocessing method sits between initial SfM and feature matching module, reducing the number of points and reference images. The feature matching module takes the SfM model constructed over the reference images ${x^r_j}$ and query image $x^q$ as inputs. It outputs the matches $\mathcal{M}=\{(p_{2D},p_{3D})\}$ the 2D-3D correspondences between 2D points in the query image and 3D points in the SfM model. A Perspective-n-Point algorithm is used to find the pose of the object with respect to the camera (represented as a transformation matrix $^{c}T_{m}$).

Different kinds of matching algorithms have been developed in the past for camera localization and pose estimation.

In \cite{hfnet_sarlin2019coarse}, the authors employed a hierarchical approach to localize a camera given an image of a landmark and an SfM model of a town. They performed nearest-neighbour searches on global descriptors to identify the closest reference images in the SfM, successfully estimating the camera's pose. Additionally, they introduced HFNet, a distilled model that generates both local and global descriptors.

Another approach, OnePose, a neural network-based direct matching method, was proposed in \cite{sun2022onepose}. This method takes as input keypoint descriptors from a query image and 3D keypoint descriptors, initialized by averaging descriptors of correspondence graph leaf nodes to produce a set of matches for object localization.

Further building on this concept, \cite{he2022oneposeplusplus} introduced OnePose++, a keypoint-free direct matching method based on the detector-free feature matcher LoFTR \cite{sun2021loftr}. OnePose++ improves performance on low-textured data but requires ground truth poses for SfM construction.

%% file: 3_dominating_set.tex
\subsection{Dominating Set}
\label{sec:dominating_set}
In graph theory, a dominating set $D$ of a graph $G$ is the set of vertices in which each vertex of $G$ is either in $D$ or a neighbour of a vertex in $D$. In simpler terms, if you consider each vertex in $D$ to have a ``reach" or ``influence" over its neighbours, then every vertex in the graph is either in $D$ or is ``influenced" by a vertex in $D$.

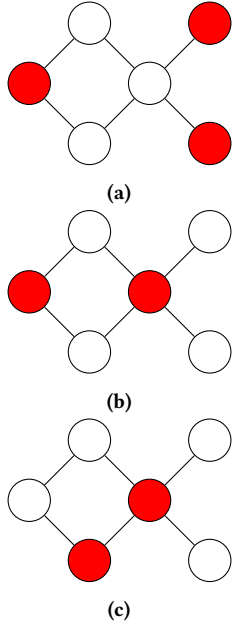
\begin{figure}[H]
\centering

\begin{subfigure}[b]{0.3\textwidth}
    \centering
    \begin{tikzpicture}[scale=0.8]
    \tikzstyle{dominating} = [circle, fill=red,draw, minimum size=16pt, inner sep=0pt]
    
    \tikzstyle{normal} = [circle, fill=white,draw, minimum size=16pt, inner sep=0pt]
    
    \node[dominating] (v1) at (0,0) {};
    \node[normal] (v2) at (1,1) {};
    \node[normal] (v3) at (1,-1) {};
    \node[normal] (v4) at (2,0) {};
    \node[dominating] (v5) at (3,1) {};
    \node[dominating] (v6) at (3,-1) {};

    \draw (v1) -- (v2);
    \draw (v1) -- (v3);
    \draw (v2) -- (v4);
    \draw (v3) -- (v4);
    \draw (v4) -- (v5);
    \draw (v4) -- (v6);
    
\end{tikzpicture}
    \caption{}
    \label{fig:sub1}
\end{subfigure}\hfill\begin{subfigure}[b]{0.3\textwidth}
    \centering
    \begin{tikzpicture}[scale=0.8]
    \tikzstyle{dominating} = [circle, fill=red,draw, minimum size=16pt, inner sep=0pt]
    
    \tikzstyle{normal} = [circle, fill=white,draw, minimum size=16pt, inner sep=0pt]
    
    \node[dominating] (v1) at (0,0) {};
    \node[normal] (v2) at (1,1) {};
    \node[normal] (v3) at (1,-1) {};
    \node[dominating] (v4) at (2,0) {};
    \node[normal] (v5) at (3,1) {};
    \node[normal] (v6) at (3,-1) {};

    \draw (v1) -- (v2);
    \draw (v1) -- (v3);
    \draw (v2) -- (v4);
    \draw (v3) -- (v4);
    \draw (v4) -- (v5);
    \draw (v4) -- (v6);
    
\end{tikzpicture}
    \caption{}
    \label{fig:sub2}
\end{subfigure}\hfill\begin{subfigure}[b]{0.3\textwidth}
    \centering
    \begin{tikzpicture}[scale=0.8]
    \tikzstyle{dominating} = [circle, fill=red,draw, minimum size=16pt, inner sep=0pt]
    
    \tikzstyle{normal} = [circle, fill=white,draw, minimum size=16pt, inner sep=0pt]
    
    \node[normal] (v1) at (0,0) {};
    \node[normal] (v2) at (1,1) {};
    \node[dominating] (v3) at (1,-1) {};
    \node[dominating] (v4) at (2,0) {};
    \node[normal] (v5) at (3,1) {};
    \node[normal] (v6) at (3,-1) {};

    \draw (v1) -- (v2);
    \draw (v1) -- (v3);
    \draw (v2) -- (v4);
    \draw (v3) -- (v4);
    \draw (v4) -- (v5);
    \draw (v4) -- (v6);
    
\end{tikzpicture}
    \caption{}
    \label{fig:sub3}
\end{subfigure}

\caption[Dominating Set]{Illustration of a graph with a dominating set. The red node represents the elements in a dominating set. Figures ~\ref{fig:sub2} and \ref{fig:sub3} represent two different minimum dominating sets for the same graph.}
    \label{fig:dominating_set}
\end{figure}

Figure~\ref{fig:dominating_set} represents the dominating set in an undirected graph. A graph can have multiple valid dominating sets. The dominating sets with the minimum number of nodes are called minimum dominating sets.

The problem of finding a minimum dominating set is classified as NP-hard \cite{Guha1998}. An approximation approach involves repeatedly executing a randomized algorithm to generate dominating sets, ultimately selecting the smallest set obtained. In this context, we employ a greedy algorithm for dominating set computation (refer to Algorithm ~\ref{alg:greedy_dominating_set})\cite{esfahanian_connectivity}. Given a graph $G=(V,E)$, with $V$ representing vertices and $E$ representing edges, the algorithm yields a dominating set $D$. While $D$ is not guaranteed to be minimum, the smallest dominating set is determined after several iterations.

\begin{algorithm}
\caption{Greedy Dominating Set}
\begin{algorithmic}[1]
\Function{DominatingSet}{$G=(V,E)$}
    \State $D \gets \emptyset$ \Comment{Initialize dominating set}
    \State $X \gets V$ \Comment {$X$ is the remaining nodes}
    \While{$X \neq \emptyset$}
        \State $u \gets \Call{RandomNode}{X}$
        \State $D \gets D \cup \{u\}$
        
        \State $X \gets X \setminus \Call{OutNeighbours}{u}$
        \State $X \gets X \setminus \{u\}$
    \EndWhile
    \State \Return $D$
\EndFunction
\end{algorithmic}
\label{alg:greedy_dominating_set}
\end{algorithm}

The concept of dominating sets extends far beyond theoretical interest, playing a critical role in various practical applications. In network design, for example, minimum dominating sets can identify the smallest number of nodes that must be equipped with transmitters to ensure communication coverage across the entire network\cite{Wan2002Distributed,Zou2011DominatingSetProblemsWSN}.

Resource allocation problems also benefit from the identification of minimum dominating sets. In scenarios where resources or services must be distributed across a network, such as distributing aid in a logistics chain or assigning service stations in urban planning, dominating sets provide a framework for ensuring that all nodes (e.g., recipients, locations) are adequately served with the fewest resource points.

Social network analysis utilizes dominating sets to identify key influencers or nodes that, if influenced or informed, can subsequently spread information to the entire network\cite{Wang2009PIDS}.

In mobile robotics, dominating sets are used for efficient robot deployment \cite{Razafimandimby2019NeuroDominatingSet}. In \cite{chand2023run}, the authors discuss how to use mobile robots to find the dominating set of an anonymous graph. In \cite{BOOIJ20091225}, a connected dominating set is used for loop closure in SLAM.

In computer vision, some of the applications of dominating sets are image segmentation, image retrieval and person re-identification \cite{tesfaye2020constrained}.

Previous works that combined Structure-from-Motion (SfM) with the concept of dominating sets primarily focused on the construction of SfM point clouds \cite{havlena2010sfm, Jiang_2022}. For instance, in \cite{havlena2010sfm}, redundant input images for constructing SfM models were removed based on image similarity, utilizing a bag-of-words approach to determine this similarity.

Our work, however, differs significantly from these approaches. Instead of focusing on the construction phase, we optimize already existing SfM models. Moreover, our method is agnostic to the specific SfM-based pose estimation technique employed. We create edges between reference images by determining if the pose of one reference image can be estimated by another within a specified threshold. More details on our approach can be found in Section \ref{sec:sfm_with_dominating_sets}. To the best of our knowledge, we are the first to use dominating sets for optimizing already generated SfM models specifically for the purpose of pose estimation

%% file: 4_method.tex
\section{Accelerating SfM-based pose estimation with dominating set}
\label{sec:sfm_with_dominating_sets}

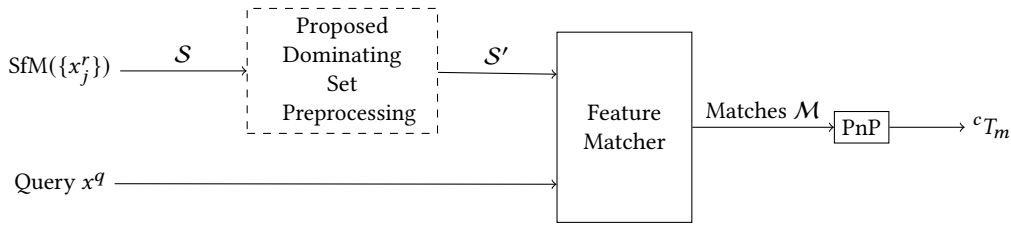
\begin{figure*}[ht]
    \centering
    \begin{tikzpicture}[scale=0.75]
        
        \node (sfm) at (-4,1) {$\operatorname{SfM}(\{x^r_j\})$};
        
        \node (frame) at (-4, -1) {Query $x^q$};
        
        \node[draw,dashed, text width=5em, minimum height=2em, minimum width=8em,text centered]  (proposed) at (1,1) {Proposed\\Dominating Set\\Preprocessing};
        
        \node[draw,text width=5em, minimum height=8em, minimum width=5em,text centered] at(6,0) (matcher){Feature\\Matcher};

        \path (matcher.0)+(3,0) node (pnp)[draw] {PnP};

        \node(t_m_to_c)[right=of pnp]{$^{c}T_{m}$};

        \draw[->] (matcher) -- node[anchor=south]{Matches $\mathcal{M}$}(pnp);
        \draw[->] (pnp) -- (t_m_to_c);

        \path [draw, ->] (frame) -- (matcher.220);
        \path [draw, ->] (sfm) -- node[above]{$\mathcal{S}$} (proposed);
        \path [draw, ->] (proposed) -- node[above]{$\mathcal{S'}$}(matcher.142);

    \end{tikzpicture}
    
    \caption{A generalized SfM-based pose estimation pipeline. Our preprocessing method sits between initial SfM and matcher, reducing the number of points and reference images.}
    \label{fig:sfm_pose_estimation}
\end{figure*}

To effectively apply the concept of a dominating set, which in graph theory refers to a subset of nodes such that every node in the graph is either in this subset or adjacent to at least one node in this subset, we first model SfM as a graph. In this graph, each node represents a reference image. A directed edge from reference image $u$ to $v$ signifies that $u$ can be used to accurately estimate the pose of $v$, considering $v$ as a query image, within a specific localization error threshold. This threshold, quantified as the ratio of the localization error bounding box coordinates to the longest diagonal, is set to 0.05 for our experiments. Algorithm ~\ref{alg:graph_generation} encapsulates the graph generation.

\begin{figure}[H]
    \centering
    \includegraphics[width=\columnwidth]{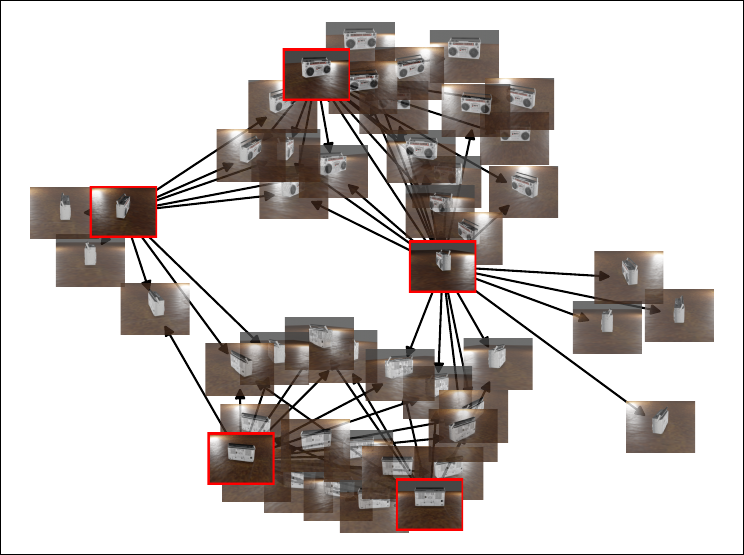}
    \caption[Visualization of the dominating set of SfM]{Visualization of the dominating set of SfM representation\footnotemark. Each node represents a reference image in the SfM representation. The dominating set is highlighted by nodes with a red border. Directed arrows signify one reference node's capability to detect another node's pose accurately. For clarity, only edges originating from the dominating set are displayed. A force-directed layout algorithm is used, which clusters the reference images close to the nearest dominating image based on localization accuracy. We can see notable similarities in images in each cluster.}
    \label{fig:dominating_set_synthetic_data}
\end{figure}

Figure~\ref{fig:dominating_set_synthetic_data} provides a visualization of this concept using synthetic data. The figure illustrates how the dominating set is formed within the SfM graph representation, highlighting the nodes in the dominating set and the directed edges that demonstrate pose estimation capability.

\footnotetext{SfM constructed using a synthetic scene. The model can be found at https://skfb.ly/oIH7J}

Finding the minimum dominating set in a graph is a complex problem for which no algorithm exists that can solve it in polynomial time. To address this challenge, we employed a randomized greedy algorithm for dominating sets, as detailed in Algorithm~\ref{alg:greedy_dominating_set}. By repeatedly applying this algorithm, we generated several dominating sets and selected the smallest among them as our minimal dominating set.

Let the given SfM model be,
\begin{equation}
    \mathcal{S} = (P^R,X^R)
\end{equation}
Where $P^R=\{p_i^r\}$ the set of points in the SfM point cloud and $X^R=\{x_j^r\}$ the reference images in SfM.

Suppose we estimated Dominating Set as per algorithm~\ref{alg:greedy_dominating_set},
\begin{equation}
    D = \{ x_k^r\}
\end{equation}

The optimized SfM is given by,
\begin{equation}
    \mathcal{S'} = (P^{'R}, X^{'R})
\end{equation}

Where,
\begin{align}
    P^{'R} &= \{p_i^r: \operatorname{parents}(p_i^r) \cap D \neq \phi\} \\
    X^{'R} &= D
\end{align}

The function $\operatorname{parents}$ gives the reference images observing the input 3D point.

The minimum dominating set can potentially improve the speed of SfM-based pose estimation since we can use the dominating set to reduce the number of points in the SfM point cloud. That means fewer points to match, which will boost the speed.

\begin{algorithm}
\caption{Construct Graph from Reference Images}
\begin{algorithmic}[1]
\State $G = (V, E) \gets (\{\}, \{\})$
\For{$x_i \in \text{ref\_imgs}$}
    \State $V.\text{append}(x_i)$
    \For{$x_j \in \text{ref\_imgs}$}
        \If{$x_i == x_j$}
            \State \textbf{continue}
        \EndIf
        \If{$\text{error}(\text{estimate\_pose}(\text{ref}=x_i, \text{query}=x_j)) < 0.05d$}
            \State $E.\text{append}((x_i, x_j))$
        \EndIf
    \EndFor
\EndFor
\end{algorithmic}
\label{alg:graph_generation}
\end{algorithm}

See figure \ref{fig:effect_of_dominating_set_on_sfm} to see the effect of applying dominating set on an SfM model. The number of reference images (represented by camera frustums) and points in the point cloud are reduced.

\begin{teaserfigure}
    \centering
    \includegraphics[width=\textwidth]{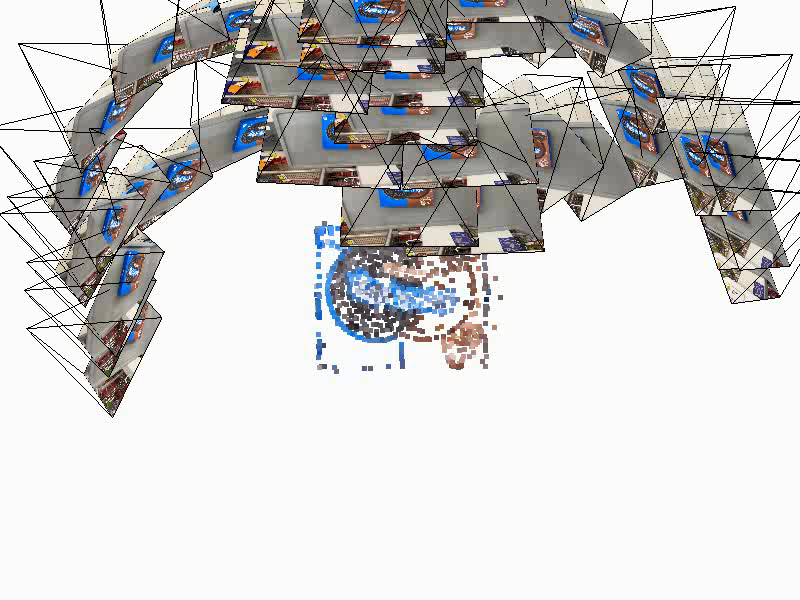}
    \includegraphics[width=\textwidth]{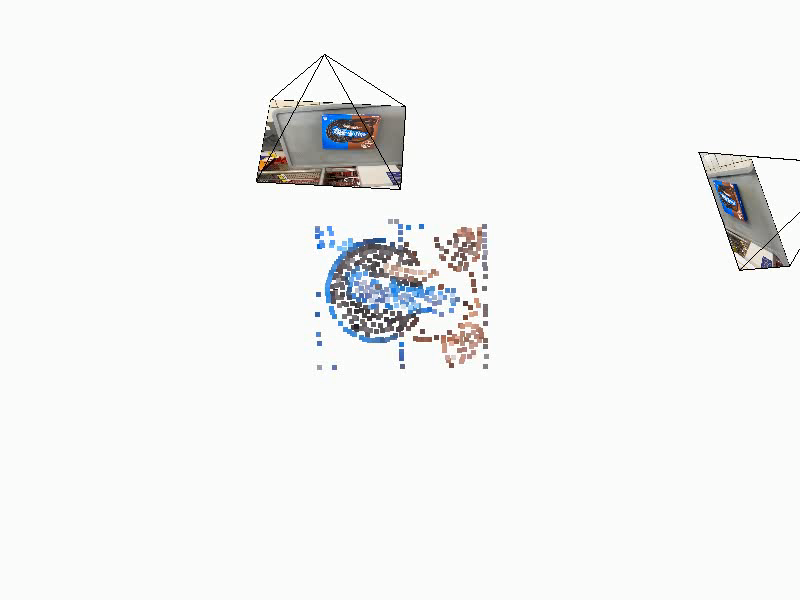}
    \caption{This figure shows the reduction of reference images and points in SfM model. Each reference image is represented as a camera frustum. The orientation and location of each camera frustum corresponds to the pose of the camera which took the reference image. (a) The SfM point cloud before applying the dominating set. (b) The SfM point cloud after applying the dominating set. The number of reference images are reduced to two and number of points are reduced but in a way they still contain the most important information for object pose estimation.}
    \label{fig:effect_of_dominating_set_on_sfm}
\end{teaserfigure}

%% file: 5_experiments.tex
\section{Experiments and Results}
\label{sec:experiments}

\subsection{Methods}
Brief explanation of each method for evaluation is given as follows, we follow the basic pose estimation pipeline shown in figure~\ref{fig:sfm_pose_estimation}. Only the feature matching module is different in each method.

\textbf{OnePose:} In OnePose, 2D feature descriptors in the query are matched with 3D feature descriptors corresponding to the points in the SfM representation. SuperPoint feature extractor is used for extracting the feature descriptors for both SfM construction and query image. 3D descriptors are calculated by averaging the 2D descriptors in each reference SfM image that observe the same point.

The feature matching is done by Graph Attention Networks (GATs) \cite{sun2022onepose} as proposed in the paper. The output of the network is 2D-3D correspondence between 2D feature points in the query image and 3D points in the SfM representation.
Then PnP\cite{lepetit2009epnp} algorithm is used to find the pose of the object as seen in the query image.

\textbf{OnePose++:} OnePose++\cite{he2022oneposeplusplus} is a similar 2D-3D matching method. Here, LoFTR\cite{sun2021loftr}, a detector-free local feature matcher, is used for finding dense correspondences. Using the dense correspondences, a coarse SfM representation is constructed. This coarse representation is then refined to construct a dense SfM representation which will be used for matching with 2D points. The OnePose++ matcher receives the SfM pointcloud and query image as inputs and outputs 2D-3D matches. Then PnP algorithm is used to find the pose of the object.

\textbf{Exhaustive Matching:} This approach involves comparing the feature descriptors from each reference image against those in the query image. It's a thorough but computationally intensive method.

The 3D points in SfM corresponding to the matched points in reference image are used as object points for the PnP algorithm and corresponding 2D points in query image are used as image points in the PnP algorithm.

For exhaustive matching, we used SuperPoint\cite{detone2018superpoint} and either SuperGlue\cite{sarlin2020superglue} or LightGlue\cite{lindenberger2023lightglue} as matcher.

\textbf{Preprocessing by Dominating Set} This is agnostic to the matching algorithm used. In this method, the number of reference images or points in the pointcloud are reduced or filtered by the most important reference images.

\textbf{Random Sampling:} Our hypothesis is that the SfM points observed in the dominating set perform better than those observed in a random sample of reference images. To validate this, we sample the same number of reference images as that of the dominating set for each object. Then, we filter SfM points and reference images. 

As evident from the results that follow, the dominating set gives better results for the same number of reference images in the random sample.

All our experiments were conducted on an NVIDIA A6000 GPU.

\subsection{Dataset}
We conducted an evaluation of the Dominating Set approach and other baseline pose estimation methods using the test split of the OnePose dataset\cite{sun2022onepose}, which comprises 80 distinct objects and 301 videos for validation purposes.

We don't use all 301 videos. Following the work \cite{sun2022onepose}, for constructing the Structure from Motion (SfM), we utilized the first video of each object, while the last video was employed for validation. The reference images are sampled by taking two frames from each 1-second interval of the first video.

\subsection{Metrics}
The metrics we chose for our experiments are \textbf{n deg - n cm}\cite{Shotton_CameraRelocalization_RGBD_2013,sun2022onepose} and \textbf{ADD-0.1d}\cite{Hinterstoier2012ModelBT}.

\subsubsection{n deg - n cm}
    
If the object localization error is within $n$ cm and the geodesic error is within $n$ degrees, then the pose estimation is considered successful.

The localization error $E_{loc}$ is the Euclidean distance between the ground truth object location ($\mathbf{t_{gt}}$) and the estimated object location ($\mathbf{\hat{t}}$) with respect to the center of the camera, see (\ref{eqn:loc_error}).

\begin{equation}
    E_{loc} = \| \mathbf{t_{gt}} - \mathbf{\hat{t}} \|_2.
    \label{eqn:loc_error}
\end{equation}

The geodesic error is the angular distance estimated orientation and ground truth orientation. 
Further, let $R_{gt}$ be the rotation matrix representing ground truth orientation and $\hat{R}$ be the rotation matrix representing estimated orientation.
The relative orientation is given by Equation~(\ref{eqn:relative_orientation}), i.e., 

\begin{equation}
    \Delta R = R_{gt}^{-1} \hat{R}.
    \label{eqn:relative_orientation}
\end{equation}
From the relative orientation, geodesic error $E_\theta$ is calculated by Equation~(\ref{eqn:geodesic_error}).

\begin{equation}
    E_\theta = \arccos\left(\frac{\text{trace}(\Delta R) - 1}{2}\right).
    \label{eqn:geodesic_error}
\end{equation}
The success rate is reported. But a main drawback of this metric is it is misleading while comparing pose estimation accuracy of objects with different sizes. Naturally a large object might produce large `n deg - n cm' error. Therefore we need to resort to a metric which takes the object size into consideration as described in \ref{sec:mod_add_01d}.

\begin{figure}
    \centering
    \includegraphics[width=\columnwidth]{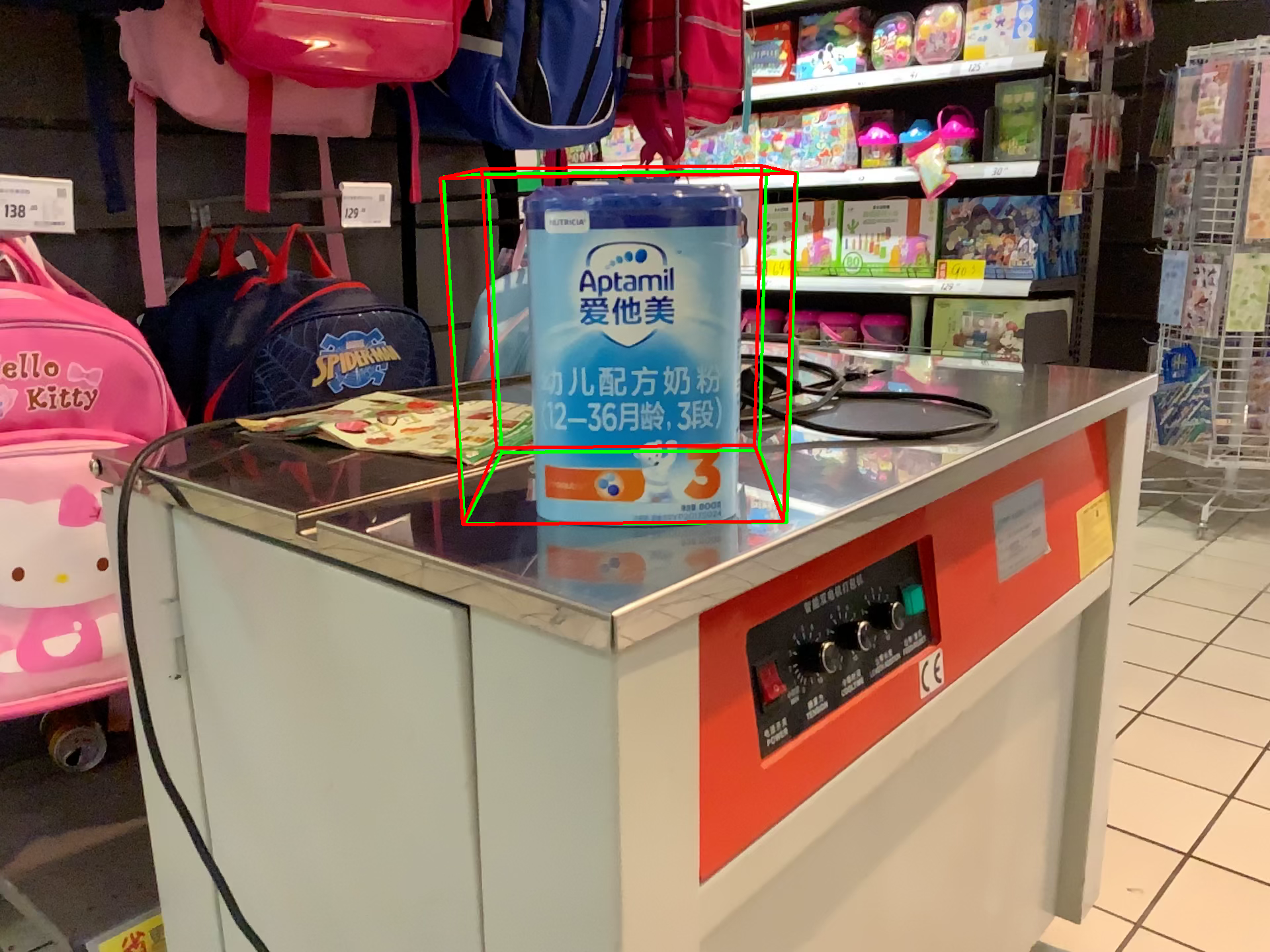}
    \caption[Bounding box visualization]{A frame in which the detected pose is visualized as a 3D bounding box. In this frame, the bounding box (red) corresponds to the predicted pose and overlaps with the ground truth bounding box (green). The matching method demonstrated here is exhaustive matching with SuperPoint as the feature extractor, Light Glue as the feature matcher and reference images taken from the dominating set.}
    \label{fig:bb_visualization}
\end{figure}

\begin{table*}[!t]
    \caption{Evaluation on OnePose Dataset \\ The midrule seperates each method. The variant is given in the square brackets.}
    \centering
    {\footnotesize \begin{tabular}{lrrrrr}
    \toprule
     Method & 1 deg - 1 cm & 3 deg - 3 cm & 5 deg - 5 cm & ADD-0.1d & FPS\\
    \midrule
    OnePose & 50.71 & 77.85 & 83.94 & 87.83 & {27.39} \\
    OnePose[Dominating Set] & 50.67 & 73.18 & 78.67 & 82.07 & 40.67\\
    OnePose[Random Sampling] & 45.04 & 64.47 & 68.83 & 72.34 & 39.49\\
    \midrule
    OnePose++ & 51.49 & 78.76 & 85.25 & 89.63 & {12.75} \\
    OnePose++[Dominating Set] & 46.45 & 72.50 & 79.51 & 83.78 & {19.12} \\
    OnePose++[Random Sampling] & 38.41 & 60.63 & 66.49 & 70.97 & {18.00} \\
    \midrule
    Exhaustive[SuperPoint-SuperGlue] & 61.94 & 85.19 & 90.76 & 94.50 & 1.40\\
    Exhaustive[SuperPoint-SuperGlue, Dominating Set] & 54.34 & 76.64 & 82.32 & 86.15 & 6.66\\
    Exhaustive[SuperPoint-SuperGlue, Random Sampling] & 45.20 & 62.92 & 67.36 & 71.23 & 6.60\\
    \midrule
    Exhaustive[SuperPoint-LightGlue] & 61.05 & 85.05 & 90.35 & 94.24 & 1.81\\
    Exhaustive[SuperPoint-LightGlue, Dominating Set] & 51.59 & 74.07 & 79.62 & 83.46 & 26.20\\
    Exhaustive[SuperPoint-LightGlue, Random Sampling] & 42.45 & 59.99 & 64.31 & 68.07 & 25.19\\
    \bottomrule
    \end{tabular}}
    \label{tab:evaluation_on_onepose_dataset_avg_by_frame}
    \end{table*}
\subsubsection{Modified ADD-0.1d}
\label{sec:mod_add_01d}

The detection is considered successful if the average localization error of points in the 3D model is below 10\%  of the diameter. Since the 3D model is unavailable, we modify the ADD-0.1d metric as the mean localization error of bounding box coordinates, following the approach used by \cite{liu2022gen6d} on PVNet dataset. The diameter is defined as the longest diagonal of the bounding box. 
The success rate is reported.

\subsection{Evaluation on OnePose Dataset}

We construct the reference image graph using Algorithm \ref{alg:graph_generation}. Each edge in the graph is determined by thresholding the pose estimated between pairs of reference images. This is achieved by matching the 2D keypoints of image $x_j$ with the 3D points corresponding to the 2D keypoints of image $x_i$. We use SuperPoint\cite{detone2018superpoint} as the 2D feature extractor and SuperGlue\cite{sarlin2020superglue} as the 2D feature matcher.

To find the minimal dominating set, we apply Algorithm \ref{alg:greedy_dominating_set} over multiple iterations, selecting 1000 iterations for optimal results. Once the most important reference images for pose estimation—the dominating set—are identified, we apply this dominating set across all pose estimation methods for evaluation.

The evaluation results for the entire OnePose test split are presented in Table~\ref{tab:evaluation_on_onepose_dataset_avg_by_frame}.  Bounding box visualization of the detected pose is given in Figure ~\ref{fig:bb_visualization}. In all the cases, the dominating set approach improves the FPS with a small performance cost. On the other hand, even though a random number of reference images increases the FPS,  it fails to reach the performance of the dominating set.

The experiments demonstrate that while the Dominating Set approach may not yield the highest accuracy, its significantly higher FPS values indicate a substantial improvement in processing speed. This balance between accuracy and speed is crucial in real-time applications, where quick decision-making based on pose estimation is necessary.

In our study, dominating set sizes in image sequences ranged from 1 to 35 images, with a mean size of 3.9125. Larger dominating sets were primarily attributed to errors in ground truth annotations of certain sequences. Therefore, improving the accuracy of ground truth annotations is expected to enhance the results.

The mean ratio between the dominating set and the number of reference images in the unfiltered SfM model is 0.057 and the median is 0.043. That is a reduction of reference images by a factor of 17 to 23. 

We examined the ratio of Structure-from-Motion (SfM) points filtered by a dominating set compared to unfiltered points. For OnePose, the mean and median ratios were around 0.44, resulting in a reduction factor of 2.27. In the OnePose++ scenario, the mean ratio was 0.26, and the median was 0.25, leading to a fourfold reduction.

\subsection{Limitations}
For better results, the training images for SfM construction should cover all the regions of the object. Suppose training images cover only one hemisphere of the object; the dominating set might remove images on the boundary of that hemisphere. An example situation can be seen in \ref{fig:effect_of_dominating_set_on_sfm}. Reference images in the boundary are removed since the reference images in the middle were able to estimate the pose of the object accurately. The accuracy will go low if the query image contains the region that overlaps only with the images removed.

%% file: 6_conclusion.tex
\section{Conclusion}
\label{sec:conclusion}
In this paper, we discussed improving the speed of Structure from Motion (SfM)-based object pose estimation with the help of dominating sets. Our findings indicate that eliminating redundant images can accelerate inference speed. To achieve this, we represented SfM as a graph and identified an approximately minimum dominating set of reference images, leading to an improved speed-accuracy trade-off.

Moreover, this method significantly decreased the needed number of reference images by a factor of between 17 and 23 and reduced the number of points required for OnePose and OnePose++ by factors ranging from 2.27 to 4.

For future work, we can broaden this application of dominating sets across additional Structure-from-Motion (SfM)-based pose estimation techniques. Although the reduction of reference images results in a minor decrease in accuracy, integrating a tracking module that leverages history of pose estimation and relevant state changes could improve accuracy without compromising real-time performance.